%% file: multiwiki_demo_paper.tex
\documentclass{sig-alternate-05-2015}

\usepackage{hyperref}
\usepackage{todonotes}
\usepackage{mathtools}
\usepackage{nicefrac}
\usepackage{subfigure}
\usepackage{stfloats}

\let\ab\allowbreak

\begin{document}

\CopyrightYear{2016}
\setcopyright{acmlicensed}
\conferenceinfo{SIGIR '16,}{July 17 - 21, 2016, Pisa, Italy}
\isbn{978-1-4503-4069-4/16/07}\acmPrice{\$15.00}
\doi{http://dx.doi.org/10.1145/2911451.2911472}


\clubpenalty=10000 
\widowpenalty = 10000





%

\title{Analysing Temporal Evolution of\\Interlingual Wikipedia Article Pairs}
%
%
%
%
%

 \numberofauthors{2} 
 \author{
 \alignauthor
 Simon Gottschalk\\
        \affaddr{L3S Research Center}\\
       \affaddr{Hannover, Germany}\\
        \email{gottschalk@L3S.de}
 \alignauthor
 Elena Demidova\\
        \affaddr{Web and Internet Science Group, ECS}\\
        \affaddr{University of Southampton, UK}\\
        \email{e.demidova@soton.ac.uk}
 }

\maketitle
\begin{abstract}

Wikipedia articles representing an entity or a topic in different language
editions evolve independently within the scope of the
language-specific user communities. This can lead to different
points of views reflected in the articles, as
well as complementary and inconsistent information.
An analysis of how the information is propagated across the Wikipedia language
editions can provide important insights in the article evolution along the
temporal and cultural dimensions and support quality control.
To facilitate such analysis, 
we present MultiWiki -- a novel web-based user interface that provides an
overview of the similarities and differences across the article pairs originating from
different language editions on a timeline.
%
MultiWiki enables users to observe the changes in the interlingual
article similarity over time and to perform 
a detailed visual comparison of the article snapshots at a particular time
point.
%



\end{abstract}

\section{Introduction}
\input{Introduction}

\section{Features \& Interface}
\label{sec:features}
\input{features}

\section{System Architecture \& Pipeline}
\input{pipeline}

\section{Dataset}
\input{dataset}

\section{Related Work}
\input{related}

\section{Conclusions}
\input{conclusion}

\small{
\section{Acknowledgments}
This work was partially funded by the ERC
under ALEXANDRIA (ERC 339233) and 
H2020-MSCA-ITN-2014
WDAqua (64279).

\bibliographystyle{abbrv}
\small{
	\bibliography{www}
}
}

\end{document}

%% file: Introduction.tex

Wikipedia is a rich interlingual information source that often reflects cultural 
differences: Wikipedia articles describing real-world entities, topics, events and 
concepts evolve independently in different language editions, as there is no 
sophisticated support for multilingual collaboration when writing articles. 
Moreover, there is a social impact on multilingual article creation, as different 
communities of Wikipedia editors vary in their perception of the topics they are 
describing. As a result, there is a distributed evolution of Wikipedia articles 
that leads to semantic differences representing linguistic points of view on 
particular topics \cite{Massa2012}, as well as complementary or sometimes
contradictory information between articles.

The temporal evolution of Wikipedia articles plays an important role when analysing 
the interlingual development of articles: For example, multilingual users 
actively transfer information from one language edition to others \cite{Rogers2013} 
and articles may be translated to complement missing
information.
Likewise, information on events impacting both language editions at a time could 
originate from one of the language editions and be adapted from the other article 
with some delay. 

Our methods aim at supporting Wikipedia editors and readers in 
better understanding of how such changes propagate across the language
editions.
To this extent, we measure the similarity of the interlingual article snapshots
at different time points and visualise them on a timeline.

\begin{figure*}[ht!]
	\centering
	\includegraphics[width=\textwidth]{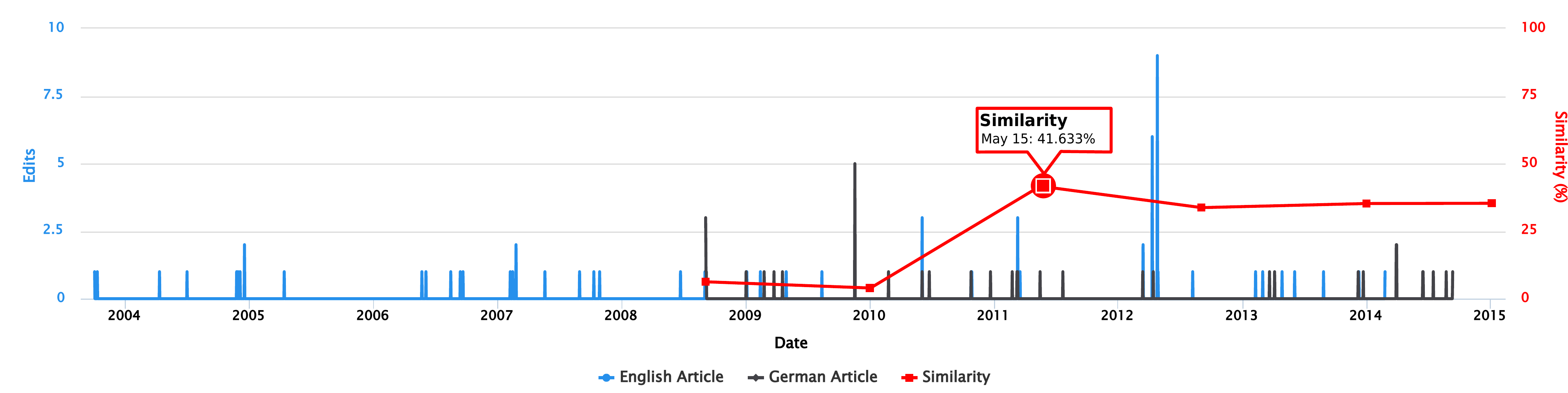}
	\label{fig:timeline}
	\caption{Timeline of the German/English article pair ``Codex Aureus of St. Emmeram''. The number of edits per article and the similarity of the article snapshots are plotted over time.}
\end{figure*}

Figure \ref{fig:timeline} gives an example of such a timeline generated by our
tool, MultiWiki, for the English/German article pair ``Codex Aureus of St.
Emmeram'', which visualises the following process: The English article was
created in October 2003, nearly five years before the German one and consisted 
of a very few sentences for a long time. Then, in 2008, the German article was
created and became immediately much longer than the English one. Hence, the
timeline shows a very small similarity within the first two years after 
the creation of the German article. Then, parts of the German texts were adapted 
to enlarge the English article, which lead to a significant increase of their
interlingual similarity. Finally, the texts evolved independently from each
other and their similarity slightly decreased.


For each particular pair of snapshots derived from the timeline, our interface 
also provides a detailed comparison. 
This comparison clearly goes beyond existing tools (e.g. \cite{Bao:2012},
\cite{Massa2012}), in particular through visualising a detailed comparison 
of the similarities and differences of the article texts.

MultiWiki presented in this demonstration is a web-based tool\footnote{The demo is publicly available 
at: \url{http://multiwiki.l3s.uni-hannover.de/demo.html}.} including the
following functionalities:

\begin{itemize}
  \itemsep-1pt
	\item Computation of the interlingual similarity between the articles in an
	interlingual article pair. The features for this similarity computation include
	the similarity of the article texts, their media elements, entities and links, 
	as well as Wikipedia editors and their locations;
	\item Tracking the changes of the interlingual similarity over time and
	visualisation of the similarity and its features on a timeline, tables and maps;
	and
	\item Visualisation of the textual similarities of an interlingual article pair
	given their snapshots.
\end{itemize} 

These functions enable MultiWiki to: 1) Present an overview of the evolution of 
the interlingual article similarity over time; and 2) Enable users to
perform a detailed article comparison at a certain time point.

%% file: features.tex


MultiWiki provides users with an overview of how the relation between two 
articles that describe the same topic in different languages changes over
time.
To this extent, MultiWiki automatically generates a timeline that visualises the
evolution of the semantic similarity
of the articles over time and the corresponding number of edits for both
articles.

In order to visualise this timeline and to enable a more fine-grained analysis 
of an article pair at a particular time point, 
we compute a semantic similarity score $Sim(A_1,A_2)$ for two article snapshots $A_1$ and
$A_2$ from different Wikipedia language editions. This score is composed of several 
similarity functions $Sim_f(A_1,A_2) \in [0,1]$ that are applied on different 
features $f \in F$.

We differentiate between the text-based features 
($F_{text}$) and features using meta information like images
and links ($F_{meta})$.
The textual similarity $Sim_{text}$ between two article snapshots $A_1$ and $A_2$ 
is computed as follows, where $w_f$ denotes the weight assigned 
to one of the similarity functions:

\begin{equation*}
Sim_{text}(A_1, A_2) = \sum_{f \in F_{text}} w_f \cdot Sim_{f}(A_1,A_2), \sum_{f
\in F_{text}} w_f = 1.
\end{equation*}

\begin{table}[b!]
	\scriptsize 
	\centering
	\begin{tabular}{|p{2cm}|p{5cm}|p{0.3cm}|}
		\hline
		\textbf{Name} & \textbf{Description} & \textbf{$w_f$} \\ \hline
		Images & Overlap of images in the articles & \nicefrac{1}{4} \\ \hline
		Wikipedia annotations & Overlap of Wikipedia annotations extracted from the articles & \nicefrac{1}{4} \\ \hline
		External links & Overlap of footnote links & \nicefrac{1}{8} \\ \hline
		External links (Hosts) & Overlap of the host names of footnote links &
		\nicefrac{1}{8} \\ \hline Editors & Overlap of editors & \nicefrac{1}{8} \\ \hline
		Editors (Locations) & Overlap of the editors' countries & \nicefrac{1}{8} \\ \hline
	\end{tabular}
	\caption{Similarity Measures based on Meta Features}
	\label{tab:featuremeasures}
\end{table}


\noindent $Sim_{meta}$, the similarity of the article snapshots computed using meta
information, is defined analogously.
We consider different feature categories, such as images, Wikipedia annotations
(i.e. references to other Wikipedia articles in the text), external links and
editors to be equally important. Therefore, we equally distribute the 
weights across these feature categories.
Tables \ref{tab:featuremeasures} and \ref{tab:textualmeasures} provide an
overview of the MultiWiki features and their weights.

The overall similarity $Sim(A_1, A_2) \in [0,1]$ of a snapshot pair 
is the weighted sum of $Sim_{text}$ and $Sim_{meta}$, where we set $\alpha=0.5$
to balance the influence of both factor types:

\begin{equation*}
Sim(A_1, A_2) = \alpha \cdot Sim_{text}(A_1,A_2) + (1-\alpha) \cdot
Sim_{meta}(A_1, A_2).
\end{equation*}

\begin{table}[b!]
	\scriptsize 
	\centering
	\begin{tabular}{|p{2cm}|p{5cm}|p{0.3cm}|}
		\hline
		\textbf{Name} & \textbf{Description} & \textbf{$w_f$} \\ \hline
		Text length & Proportion of the number of characters & \nicefrac{1}{3} \\
		\hline Text overlap & Jaccard overlap of the English texts & \nicefrac{1}{3}
		\\ \hline Text passage coverage & Percentage of characters that have been
		aligned with the other article & \nicefrac{1}{3} \\ \hline
	\end{tabular}
	\caption{Textual Similarity Measures}
	\label{tab:textualmeasures}
\end{table}


\begin{figure*}[th!]
	\centering
	\begin{subfigure}[Images]{\includegraphics[width=.31\linewidth] {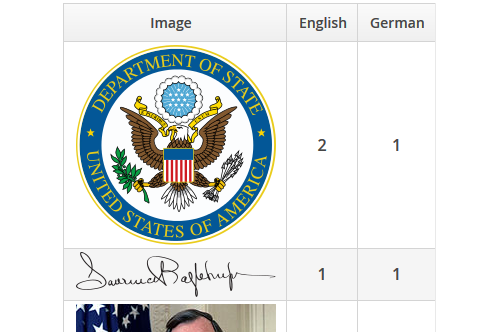}
		\label{fig:images}
		}%
	\end{subfigure}\hfill
	\begin{subfigure}[Links (Host Names)]{\includegraphics[width=.31\linewidth] {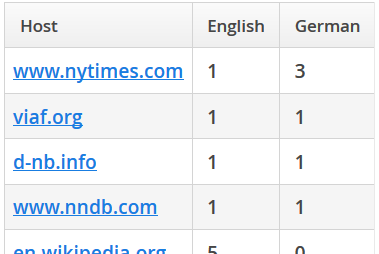}
		\label{fig:links}
		}%
	\end{subfigure}\hfill
	\begin{subfigure}[Editor Locations]{\includegraphics[width=.31\linewidth] {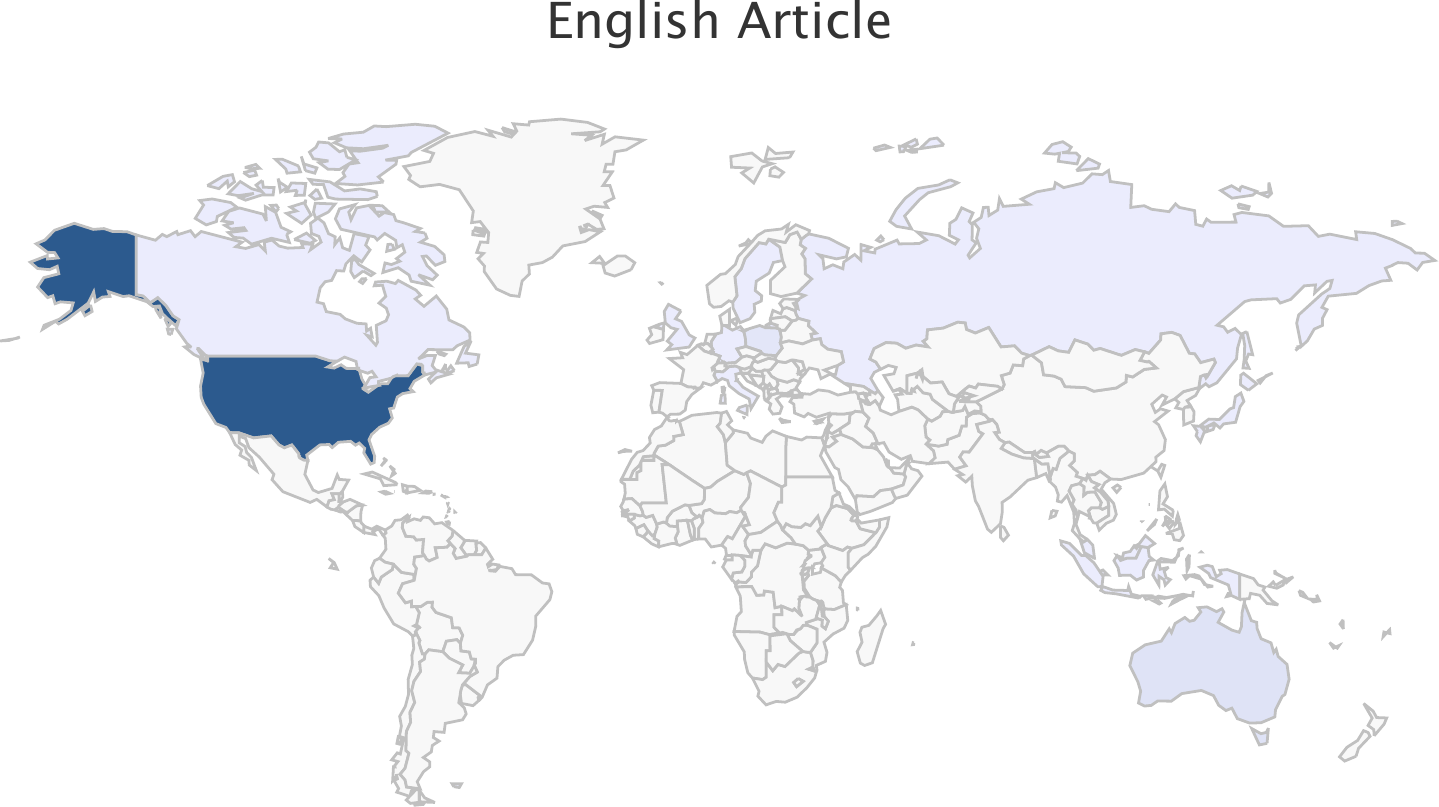}
		\label{fig:editors}
		}%
	\end{subfigure}%
	\caption{Selected Feature Similarities for the English/German article pair on ``Lawrence Eagleburger''}
	\label{fig:features}
\end{figure*}

\begin{figure*}[b]
	\centering
	\includegraphics[width=\textwidth]{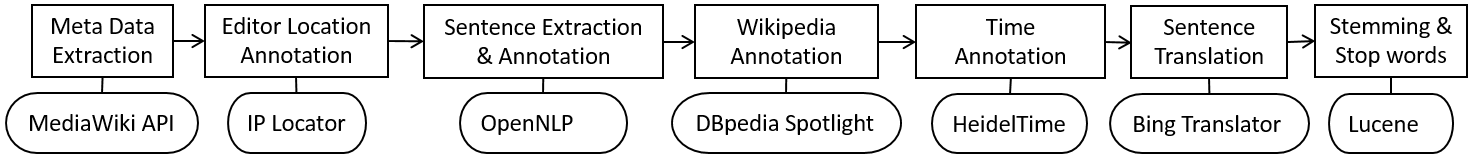}
	\caption{Processing Pipeline for a Wikipedia Article}
	\label{fig:pipelien}
\end{figure*}

In the following, we describe the features, the similarity functions based on them 
and how we present the results to the user: Our interface does not only 
show the numeric similarity scores, but also visualises the  
similarities based on particular features as illustrated in Figures
\ref{fig:features} and \ref{fig:passages}.

\subsubsection*{Images}

When comparing two Wikipedia articles, one of the main visual impressions is given by the images placed in them.
Therefore, we compute the fraction of images used in both articles. In the
interface, we oppose them in a table like the one shown in Figure \ref{fig:images}.

\subsubsection*{External Links}

Motivated by the findings in
\cite{Rogers2013}, MultiWiki aligns external links 
mentioned in the footnotes of the articles and their hosts.
%
%
As illustrated in Figure \ref{fig:links}, the host names of the external links 
are shown to the user ranked by their overlap frequency. 
Those at the
top are frequently mentioned in both articles 
and are therefore good indicators for the topical overlap of the articles.

\subsubsection*{Wikipedia Links and Annotations}

Two articles are similar if they refer to 
the same entities, concepts and topics \cite{Massa2012}. To extract
this kind of annotations, we rely on two sources: 
(1) Wikipedia links manually added to articles
by Wikipedia editors; and (2) Wikipedia 
links automatically created by applying an entity annotation
tool (DBpedia Spotlight).
While the user-defined links are precise, but rather sparse and without
repetitions, the automatic annotation raises the recall of the Wikipedia references. 
These references 
are aligned across languages
by using interlingual 
article links provided by Wikipedia. 

\subsubsection*{Wikipedia Editors}

Wikipedia articles are thoroughly based on the input of the Wikipedia 
editors. 
Some multilingual editors tend to edit articles describing the same entity in 
different languages, such that similar content is spread across languages
\cite{Rogers2013}.
MultiWiki computes the overlap of the editors 
that contributed to the articles under comparison. Furthermore, the linguistic 
point of view (as defined in \cite{Massa2012}) emphasises the cultural 
differences between language communities. Therefore, we determine the countries 
where anonymous editors come from and compute the similarity of the country 
distribution per article. As this location 
similarity \ab $Sim_{\textnormal{loc}}$ ought to measure the similarity of 
the distribution, but should not be biased by differences in the number of 
anonymous editors, it is computed by Equation \ref{eq:editor_loc} 
(where $L$ is the set of locations and each article $A_i$ is assigned a 
set of editors $e \in E_i$): Before computing the overlap, the number of 
editors is matched by multiplying $E_2$ by $\frac{|E_1|}{|E_2|}$.

\begin{equation*} \label{eq:editor_loc}
\begin{multlined}
\small
Sim_{\textnormal{loc}}(A_1,A_2)= \\
\frac
{
	\sum\limits_{l \in L} min(
	|\{e \in E_1 | e.\textnormal{loc} = l\}|,|\{e \in E_2 | e.\textnormal{loc} = l\}| \cdot \frac
	{|E_1|}
	{|E_2|}
	)
}
{
	|E_1|
}
\end{multlined}
\end{equation*}

In the MultiWiki interface, two world maps like the one presented in 
Figure \ref{fig:editors} are showing the editors' origins for a pair of article snapshots.

\subsubsection*{Textual Overlap}

\begin{figure*}[t!]
	\centering
	\includegraphics[width=0.9\textwidth]{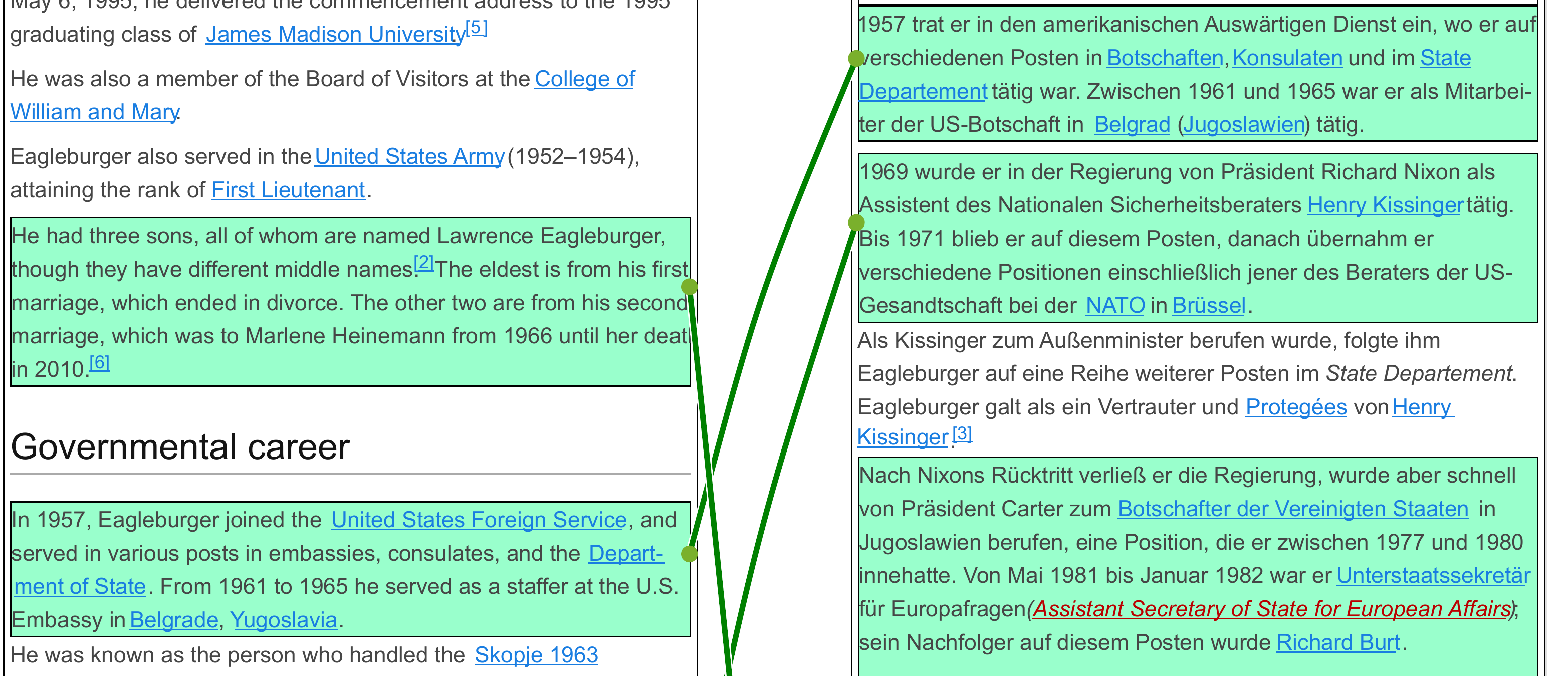}
	\caption{Text Comparison. The English article on ``Lawrence Eagleburger'' is on the left, the German one on the right. Aligned passages are marked green and connected with matching passages in the other article.}
	\label{fig:passages}
\end{figure*}

MultiWiki 
provides a fine-grained comparison of the article texts by 
aligning semantically similar text passages in an article pair
as illustrated 
in Figure \ref{fig:passages}.

This alignment is facilitated by an bottom-up approach: At first, sentences with
overlapping information are aligned using the following features: 
(1) Cosine similarity of the vectors representing text passages 
using terms (automatically translated to English); 
(2) Similarity of the Wikipedia entities metioned in the sentences; 
(3) Time similarity using temporal expressions automatically extracted 
from the sentences. Consequently, sentence pairs whose similarity score exceeds 
a pre-determined threshold value are merged in a bottom-up manner: 
As long as the similarity of an aligned text passage pair increases, 
the text passage is expanded by adding neighboured sentences from one of the
articles or by merging two text passage pairs in a close neighbourhood. 
Given the aligned text passages, they are presented to the user by 
showing both articles side by side and connecting them as illustrated 
in Figure \ref{fig:passages}. With this visualisation approach, 
the user immediately obtaines an overview of the similarities and differences in
the articles. 

\subsubsection*{Textual Similarity}

Regarding the textual similarity score of an article 
pair \ab $Sim_{text}(A_1, A_2)$, three measures are derived from the article 
texts: 1) \textit{Text passage coverage} is the percentage of text  
aligned as part of text passage pairs; 2) \textit{Text length similarity}
compares the length of the articles; and 3) \textit{Text overlap similarity} is
the Jaccard similarity of the English (machine-translated) texts.



%% file: pipeline.tex
In order to facilitate the comparison of articles based on the proposed features 
and to analyse their temporal evolution, each article pair runs through a
preprocessing pipeline. In a first step, we need to inspect the revision 
history in both languages to create a set of article snapshots pairs that
fulfils the following condition: Both snapshots in an article pair must have been on-line 
at the same time. This ensures that dissimilarities between the articles are not 
based on temporal shifts. Besides, the selected article pairs should be evenly 
distributed over time to provide a better overview.

For each article snapshot, we collect meta information, extract
the sentences and annotate them with the features presented in Section \ref{sec:features} 
using state-of-the-art methods. 
The pipeline for extraction and annotation is depicted in Figure
\ref{fig:pipelien}.
At first, we fetch the texts of 
each article and collect its contributing editors from the revision history. 
For each of the unregistered, anonymous editors, the IP address is provided, 
which makes it possible to determine their country. To allow for the text
passage alignment, the text is split into sentences; then the features needed
for the sentence alignment are collected.
The results of the preprocessing pipeline are stored in a database. 
Additionally, the complete set of interlingual Wikipedia links 
is stored, such that the final comparison of the article pair including 
the computation of similarity values and the text passage alignment 
can be performed at run time.

%% file: dataset.tex
To facilitate the demonstration of our methods, we extracted a set of 79 randomly 
selected article pairs to be investigated using MultiWiki. 
For each article pair, we
collected eight snapshot pairs on average.
The majority of 54 article pairs is German/English, 
14 are Dutch/\ab English and the remaining 11 pairs are 
Portuguese/\ab English. As we extract multiple snapshots 
of each article to construct the timelines, a total of 637 snapshot pairs 
has been extracted and annotated. Among these article pairs, there is the 
politician ``Lawrence Eagleburger'', the German film ``Der Stern von Afrika'' 
(movie) and the ``General Post Office''.

%% file: related.tex
The multilingual Wikipedia has been a research target for different disciplines. 
One area of interest in this context is the study of differences in the language 
points of view in Wikipedia \cite{Massa2012}. 
In \cite{Rogers2013}, Rogers conducts an intensive case study where he compares the 
highly controversial articles on the Srebrenica massacre in several Wikipedia language editions. 
In his work, all of the investigations like image alignment and a
comparison of the table of contents are made manually. To simplify such tasks in future, 
some of the features mentioned in \cite{Rogers2013} have been implemented in MultiWiki.

Two of the few existing tools to support cross-lingual analytics in Wikipedia 
are Manypedia \cite{Massa2012} that provides an automatic translation to English, 
article statistics and overlap of Wikipedia links, and 
Omnipedia \cite{Bao:2012}, which can be used to compare the given Wikipedia links for
up to 25 language editions simultaneously. Another tool that helps to explore the 
temporal development of a single article is Contropedia \cite{Borra2015} 
that focusses on the interaction of Wikipedia editors.
Unlike existing tools, MultiWiki excibits unique features such as 
temporal analysis of interlingual article similarity and detailed textual
comparison of article snapshots.

%% file: conclusion.tex

In this paper we presented MultiWiki -- a novel user interface to conduct a
detailed visual analysis of the similarities and the differences across the Wikipedia 
articles on the same topic written in
different languages. 
MultiWiki provides means to explore 
how information is propagated across the language editions by observing how the
similarity of an article pair evolves over time. Moreover, it enables
a detailed visual comparison of the article snapshots based on users, media
elements, entities, and links as well as a detailed analysis of textual
similarities.